\documentclass[10pt,twocolumn,letterpaper]{article}

\usepackage{cvpr}              

\usepackage[dvipsnames]{xcolor}

\definecolor{cvprblue}{rgb}{0.21,0.49,0.74}
\usepackage{float}
\usepackage[accsupp]{axessibility}

\def\paperID{23} 
\def\confName{CVPR}
\def\confYear{2024}

\title{CMOSE: Comprehensive Multi-Modality Online Student Engagement Dataset with High-Quality Labels}

\author{
    Chi-Hsuan Wu\textsuperscript{\rm 1}, 
    Shih-yang Liu\textsuperscript{\rm 1}, 
    Xijie Huang\textsuperscript{\rm 1}, 
    Xingbo Wang\textsuperscript{\rm 1}, 
    Rong Zhang\textsuperscript{\rm 1}, 
    Luca Minciullo\textsuperscript{\rm 2} \\
    Wong Kai Yiu\textsuperscript{\rm 2}, 
    Kenny Kwan\textsuperscript{\rm 2}, 
    Kwang-Ting Cheng\textsuperscript{\rm 1} \\
    \textsuperscript{\rm 1}Hong Kong University of Science and Technology,
    \textsuperscript{\rm 2}LifeHikes\\
    {\tt\small \{cwuau, sliuau, xhuangbs, xingbo.wang, rzhangab\} @connect.ust.hk} \\
    {\tt\small \{luca.minciullo, tim.wong, kenny.kwan\} @lifehikes.com}, {\tt\small timcheng@ust.hk} \\
}

\begin{document}
\newcommand{\xijie}[1]{{\textcolor{blue}{\it Xijie: #1}}}
\maketitle
\begin{abstract}
Online learning is a rapidly growing industry. However, a major doubt about online learning is whether students are as engaged as they are in face-to-face classes. An engagement recognition system can notify the instructors about the student’s condition and improve the learning experience. Current challenges in engagement detection involve poor label quality, extreme data imbalance, and intra-class variety -- the variety of behaviors at a certain engagement level. To address these problems, we present the CMOSE dataset, which contains a large number of data from different engagement levels and high-quality labels annotated according to psychological advice. We also propose a training mechanism MocoRank to handle the intra-class variety and the ordinal pattern of different degrees of engagement classes. MocoRank outperforms prior engagement detection frameworks, achieving a $1.32\%$ increase in overall accuracy and $5.05\%$ improvement in average accuracy. Further, we demonstrate the effectiveness of multi-modality in engagement detection by combining video features with speech and audio features. The data transferability experiments also state that the proposed CMOSE dataset provides superior label quality and behavior diversity. 
\end{abstract}    
\section{Introduction}
\label{sec:intro}

Online learning has greatly drawn people’s attention in recent years. The outbreak of COVID-19 also increased the demand for online classes. However, people doubt whether online classes are as effective as face-to-face classes. Research has also shown that students often have a lower attention level in online classes \cite{Smith_Schreder_2021}. A model capable of classifying students’ engagement levels can inform the instructors to pay caution to specific participants and reflect the overall effectiveness of the online classes. 

In face-to-face classes, the instructors usually rely on interactions between the students, their emotions, facial expressions, and speech to verify the engagement level of each student \cite{walker2021student}. However, interaction features are missing in online mode because the students are muted and cannot discuss with each other most of the time. There are other challenges, such as the noisy background of each webcam and various illumination levels. Therefore, a model to automatically detect student engagement levels for online scenarios is necessary to enhance the learning outcome.

\begin{figure}[t]
\includegraphics[scale=0.42]{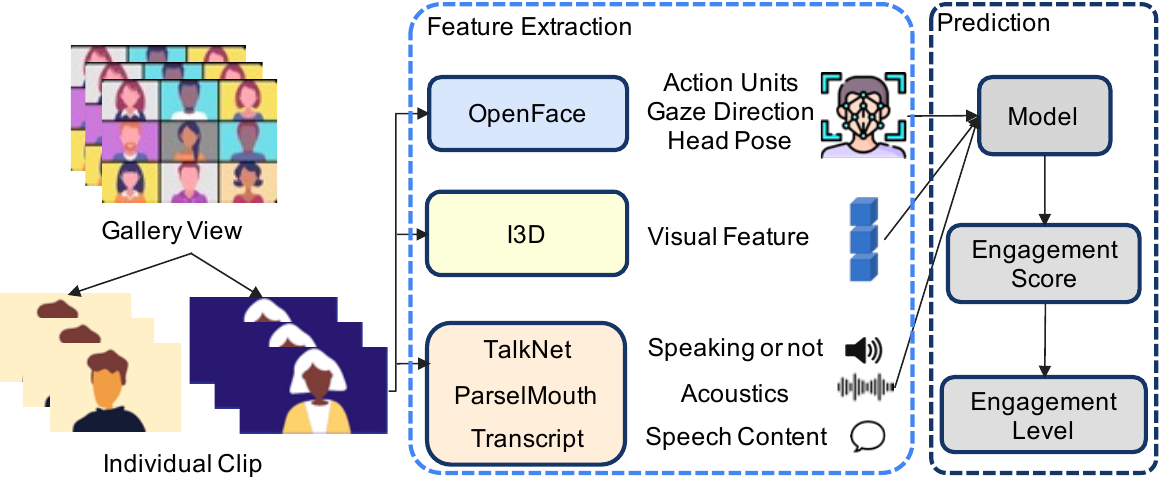}
\caption{After transferring the gallery video into individual clips, we utilized pre-trained modules to extract visual, audio, and speech features. These features are the input of the model to predict the engagement score. The engagement level is further assigned based on pre-defined thresholds. }
\label{fig:one}
\end{figure}

Existing datasets such as DAiSEE \cite{gupta2016daisee} and EngageWild \cite{kaur2018prediction} separate student engagement levels into four classes, namely highly disengaged (HD), disengaged (DE), engaged (EG), and highly engaged (HE). We follow the setting to classify the degree of engagement of each subject into four classes from the 10-second webcam video. To capture the nuance engagement difference, we follow \citet{dhall2020emotiw} to let the model output a scalar as engagement score and further assign the class based on thresholds. Figure \ref{fig:one} demonstrates the overview of our method.

Data imbalance is universal in existing datasets because students engage most of the time while only disengaging for a short period. Trained by the imbalanced data, the model may exhibit bias towards the majority class \cite{johnson2019survey}. Another characteristic is the ordinal relationship among the four degrees of engagement classes. Prior works in EmotiW2020 Challenge \cite{dhall2020emotiw} transformed class labels into scalars and used MSE Loss for training. Such transformation is inferior because the class label only indicates a range of engagement levels. For instance, some “engaged” students may appear more engaged than others with the same label. The variety of behaviors in a class, termed intra-class variance \cite{farzaneh2021facial}, shows that imposing a ground truth on data from the same class is unsuitable. To address these challenges, we introduce contrastive learning to engagement detection and propose MocoRank, a training mechanism to tackle data imbalance, intra-class variation, and ordinal relationships. 

In terms of representation learning, previous studies were limited to either frame-wise high-level features (Head Position, Gaze Direction, and Facial Action Units) from Openface \cite{baltrusaitis2018openface} or deep features from CNN architectures. However, using high-level features alone may ignore important unselected features, and using deep features alone may fail to extract the most relevant features. In contrast, our approach combines pre-trained spatial-temporal representations \cite{carreira2017quo} with high-level features to enhance the model performance. Additionally, we incorporate audio and speech features. While audio and speech have been previously used for face-to-face classroom engagement detection \cite{sumer2021multimodal}, they have not been extensively discussed in the context of online class engagement analysis. 

The label quality is another crucial factor in training an engagement detection model. Therefore, this work presents a Comprehensive Multi-modal Online Student Engagement dataset (CMOSE), with high-quality labels annotated by annotators trained by psychology experts. Extensive experiments on the engagement detection task demonstrate the outstanding quality and transferability of the CMOSE dataset. We summarize our contributions as follows: 

\begin{itemize}
    \item We present CMOSE, a comprehensive multi-modal online student engagement dataset with high-quality labels.
    \item We demonstrate the generalization ability of the CMOSE dataset by conducting transferability experiments on other engagement datasets.
    \item We propose MocoRank, a training mechanism designed to handle data imbalance, intra-class variation, and ordinal relationships for engagement prediction.
    \item  We combine different levels of visual features and audio features to enhance the performance, facilitating future research on multi-modality in engagement prediction.
\end{itemize} 

\section{Related Work}
\label{sec:relatedwork}
\subsection{Representation Learning}

Engagement prediction research has focused on high-level features, which are more interpretable by humans, or low-level features from deep neural networks. High-level features offer the advantages of noise reduction. \citet{copur2022engagement} and \citet{niu2018automatic} utilized GAP features (Gaze, Facial Action Units, Head pose) and employed temporal networks such as GRU or LSTM to model temporal information. However, high-level features have the limitation of ignoring subtle movements or informative behaviors that are not captured by the chosen features.

Recently, deep learning approaches have gained significant popularity. Hybrid design comprising a CNN architecture and a temporal network to capture spatial-temporal information becomes common. \citet{abedi2021improving} incorporated Resnet and Temporal Convolutional Network for prediction. \citet{liao2021deep} combined SENet and LSTM with global attention layers. \citet{mehta2022three} utilized 3D DenseNet with self-attention to capture global relationships among the features. \citet{ikram2023recognition} divided the video into small segments and predicted the engagement level using the learner's affective state of each segment. However, these methods are not interpretable and do not demonstrate superior results compared to GAP features.

\subsection{Ordinal Regression}

From highly disengaged to highly engaged, the four engagement classes are ordered. Previous work \cite{cao2019learning, mehta2022three, selim2022students}, predicted the probability of each class and failed to incorporate the ordinal relationship. By contrast, other work represented the engagement as a scalar \cite{dhall2020emotiw, copur2022engagement, liao2021deep}. These works assigned a numerical ground truth to each engagement class and used MSE Loss for training. However, transferring a strict numerical ground truth to describe an engagement class is inferior because the engagement class only implies a range of engagement levels. In other words, even for data from the same class, we can sometimes tell one is more engaged than the other.

\subsection{Addressing Data Imbalance Problem}

Class imbalance is the most critical feature in previous engagement detection datasets \cite{gupta2016daisee, kaur2018prediction}. Severe imbalance can lead to overfitting, impeding the model to generalize to unseen data. Prior works attempted to solve this problem by loss designs such as class-balance cross-entropy, class-balance focal loss \cite{mehta2022three}, and LDAM Loss \cite{cao2019learning}.

Deep Metric Learning (DML) is common in Facial Expression Recognition to handle data imbalance. DML is used to constrain the embedding space to obtain well-discriminated deep features and maximize the similarity between features of the same class. \citet{liao2021deep} implemented Center Loss to reduce the embeddings distance from the same class. \citet{wang2019bootstrap} designed Rank Loss to regularize the average embedding of each class and encourage the ordinal relationship. \citet{copur2022engagement} utilized Triplet Loss to separate engage and disengage features. DML has shown its importance in handling intra-class variations and inter-class similarities in engagement detection.

\subsection{Engagement Detection Datasets}

DAiSEE \cite{gupta2016daisee} and EngageWild \cite{kaur2018prediction} are the two main engagement detection datasets. Labels of the engagement datasets are often being challenged. DAiSEE relied on crowdsourcing for annotation. While unreliable annotators were filtered out, the label quality has been questioned by previous studies \cite{liao2021deep, mehta2022three}. Many works have struggled to accurately distinguish between disengaged (DE), engaged (EG), and highly engaged (HE) \cite{savchenko2022classifying}. As for EngageWild, five labelers were assigned to annotate the labels with the same guidelines on matching facial expressions with engagement levels. Though the quality of the label has significantly improved, the dataset is very small in size, which can barely represent the pattern of HD.

\section{CMOSE Dataset}

We now present the CMOSE Dataset, a collection of individual student video clips from online presentation training classes. These videos capture participants' multi-modal behavior across various in-the-wild scenarios. Each video clip is associated with an engagement label assigned by labelers who have undergone specialized training from psychologists. The dataset will be made publicly available.

\subsection{Data Collection}
The raw data comprises gallery view recordings from online presentation training classes. These classes involve one coach and multiple participants. We extract the bounding boxes of each person to separate individual videos, and each video is segmented based on time-stamped utterances. This segmentation strategy allows us to capture the fine-grained dynamics of engagement in online learning.

The subjects in different segments display a diverse range of engagement levels, accompanied by various engagement-related behaviors, such as looking down, looking away, and nodding. Additionally, since participants were encouraged to freely express their ideas, some segments feature participants speaking. This diversity in behavior and engagement levels enriches the dataset, providing researchers with valuable insights for analyzing and understanding engagement in online learning.

There are 9 training classes, involving a total of 102 participants. The participants are made up of people from different races with a male-to-female ratio of 0.65:1. Following the segmentation process, the dataset comprises a vast collection of 12,193 individual video segments, within which 2930 video segments contain speeches. The video segments were captured at 25 fps and $412 \times 234$ resolution. The length of these segments varies, with an average of 13.72 seconds. Each video segment is given a number specifying its training class and a timestamp indicating where it is located in the individual video.

\subsection{Data Annotation}
The reliability of the label in the engagement detection datasets is often challenged. Unlike DAiSEE and EngageWild, the CMOSE dataset stands out as the first engagement video dataset to incorporate labels based on the advice of psychologists. 

To ensure the reliability of the data labels, we invited three experienced teachers with rich domain knowledge to provide a list of engagement-indicating behaviors such as active head movements, looking down, etc. We provide the full list of behavior patterns and their indicated engagement score suggested by the psychology experts in Supplementary Material. Seven labelers were asked to follow the guidance to annotate the videos. The video segments were labeled into four classes, namely, highly disengaged (HD), disengaged (DE), engaged (EG), and highly engaged (HE). The Intraclass Correlation Coefficient ICC($2$,$1$) for the dataset is $0.84$ ($95$\% CI: $0.83$ to $0.85$) indicating a high level of agreement among the labelers.

The label distribution compared to DAiSEE and EngageWild is in Table \ref{tab:datadis}. The most significant advantage of the CMOSE dataset is the considerably larger number of data instances in HD and DE. This increase in the HD and DE classes expands the behavioral spectrum associated with the disengaging state. Though the CMOSE dataset contains data imbalance property, the diversity of minority classes may greatly affect the classification result \cite{4938667} and alleviate the data imbalance problem \cite{aggarwal2021minority}.

We partition the dataset randomly, allocating $70\%$ for training, $20\%$ for validation, and $10\%$ for testing purposes. Information regarding the dataset splits will be released for transparency and reproducibility.

\begin{table}[]
\fontsize{9pt}{0pt}
\centering
\begin{tabular}{l|cccc|c}
\toprule
Dataset   & HD & DE   & EG  & HE & Total\\ 
\midrule
CMOSE   & 346 & 2208  & 8469 & 1170 & 12193 \\  
EngageWild   & 9  & 45  & 100   & 43 &  197\\ 
DAiSEE  & 61   & 459  & 4477  & 4071 & 9057 \\ 
\bottomrule
\end{tabular}
\caption{Comparision of the data distribution of CMOSE, DAiSEE, and EngageWild dataset.}
\label{tab:datadis}
\end{table}

\subsection{Characteristics of the Dataset}
\begin{figure}
\centering
\includegraphics[scale=0.65]{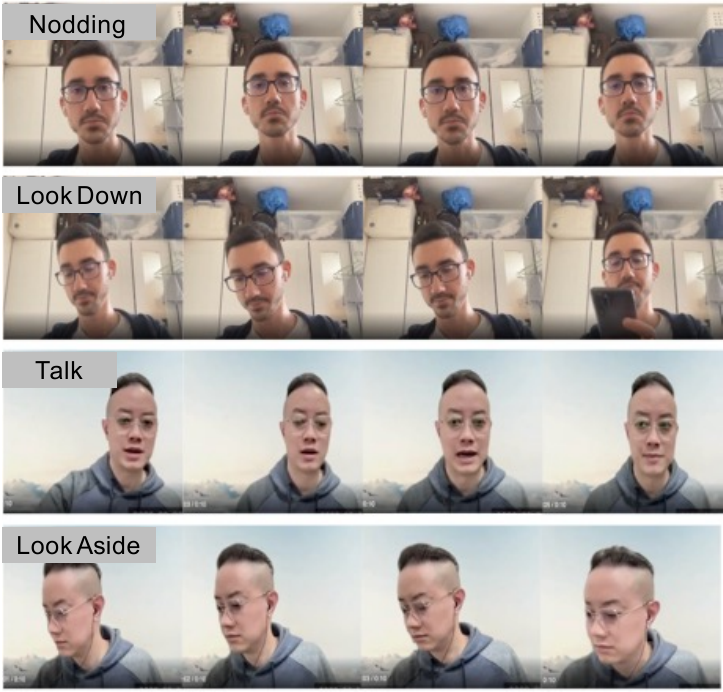}
\caption{Various behaviors included in CMOSE Dataset such as nodding, looking down, speaking, and looking away.}
\label{fig:action}
\end{figure}

The CMOSE dataset comprises subjects displaying behaviors indicative of engagement levels, as outlined by psychology experts. Various actions, including nodding, speaking, looking away, and looking down (illustrated in Figure \ref{fig:action}) are observed. In real situations, people with similar engagement levels may act differently from one another, reflecting the dataset's fidelity to genuine settings.

The ``in-the-wild'' setting can also be seen by comparing the figures in Figure \ref{fig:action}. The video of one man shows a background of a room, while the other man uses a virtual background. ``In-the-wild'' setting is important as it can reflect the real situation in online classes, where different illumination and virtual backgrounds may be shown.

CMOSE dataset also provides various modalities to facilitate future studies on incorporating different features. Apart from the visual features, we utilize TalkNet \cite{tao2021someone} to recognize the speaking subject. Speech content is also detected using the Live Transcript function in Zoom. Features related to the speech include the speech content, text length, acoustics (volume and pitch), the sentiment of the speech, etc. Additionally, information in the chatroom and the reply frequency of each participant are provided. We believe a great variety of modalities could enhance further studies on multi-modality and group engagement.

\subsection{Subject Privacy and Ethical Issue}
All participants and coaches (teachers) featured in the video segments of the CMOSE Dataset have provided informed and signed consent for the dataset to be distributed. All participants have a similar distribution of engagement levels, and we do not find specific biases toward certain participants, genders, or races after annotation.

\section{Method}

\begin{figure*}[t]
\centering
\includegraphics[width=0.95\textwidth]{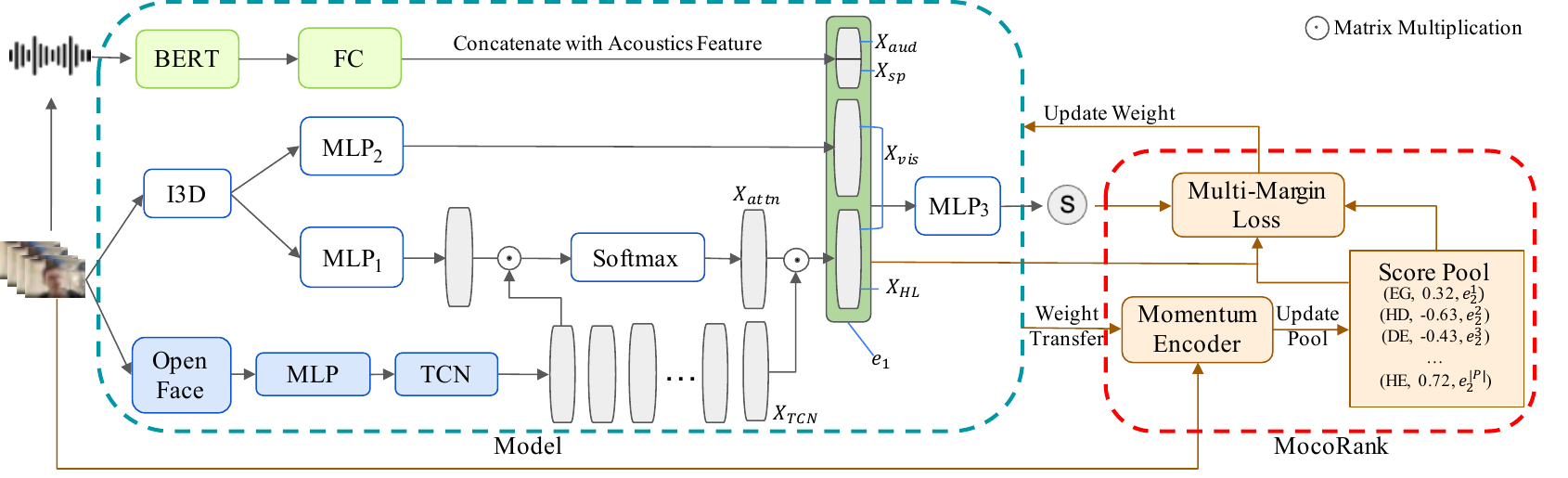}
\caption{Model structure and the training mechanism MocoRank. After the model predicts the scores for the batch of videos, the Multi-Margin Loss is calculated by comparing the scores with the triplets in the Score Pool. Next, the model will be updated and the same batch of videos will be sent to the Momentum Encoder to update the Score Pool. Lastly, parts of the weight of the model will be transferred to the weight of the Momentum Encoder. }
\label{fig:model}
\end{figure*}

In this section, we introduce our multi-modal model structure and how we train our model using the proposed MocoRank. The overview of our method is in Figure \ref{fig:model}.

\subsection{Feature Extraction}
We utilize OpenFace 2.2.0 \cite{baltrusaitis2018openface} to extract the high-level features of the subjects. These features include gaze directions, head position, and facial action units. High-level features were commonly used in previous work \cite{niu2018automatic, copur2022engagement} for detecting engagement levels. The combination of high-level features can represent engagement-related features such as nodding, yawning, looking down, etc. Also, utilizing high-level features can reduce the noise such as the video backgrounds. The details of the extracted features are as follows:

\begin{itemize}
    \item Gaze Direction and Angles: Three coordinates to describe the gaze direction of left and right eyes respectively. Two scalars to describe the horizontal and vertical gaze angles.
    \item Head Position: Three coordinates to describe the location of the head to the camera.
    \item Head Rotation: Describe the rotation of the head with pitch, yaw, and raw.
    \item Facial Action Units (AUs): Describe the intensities of 17 AUs and the presence of 18 AUs as scalars.
\end{itemize}

While the high-level features can capture the frame-wise information, some temporal information, such as body motions, may not be fully captured. Inflated 3D Network (I3D) \cite{carreira2017quo} is a widely adopted 3D video classification network that contains a 3D convolutional network and optical flow to extract spatiotemporal information. We use visual features from the I3D Network pre-trained on Kinetics 400 \cite{kay2017kinetics} to compensate for the neglected information.

For the audio feature, we utilize Parselmouth \cite{jadoul2018introducing} to extract the acoustics, which include the volume vibration and the pitch. We also use the speech content extracted by the Zoom Live Transcript as input. Multi-modality and audio features have often been used in cognitive recognition, such as depression recognition \cite{niu2020multimodal}. However, a few works have considered audio features in engagement detection. We provide audio features to encourage future studies on multi-modality engagement detection. Further details of high-level features, visual features, audio, and speech features are provided in Supplementary Material.


\subsection{Model Structure}
\subsubsection{High-level Features and Temporal Convolutional Network}
Inspired by \citet{copur2022engagement}, a video is represented as a sequence of $D$ dimension high-level features, and we separate the sequence into $T$ chunks with equal lengths. For videos under 10 seconds, we repeat the video until it contains more than $250$ frames. Further, we derive the minimum, maximum, and variance of each feature within each chunk and concatenate them into $ps \in \mathbb{R}^{3D \times T}$.

Next, we utilize a Temporal Convolutional Network (TCN) \cite{bai2018empirical} to capture temporal patterns from $ps$. The output from TCN is denoted as $X_{TCN} \in \mathbb{R}^{C \times T}$, where $C$ is the dimension of the hidden layers. While TCN may not be as intricate as Transformer-based models, our empirical study shows that TCN surpasses models like Bi-LSTM and Vanilla Transformer for our prediction tasks which are haunted by the data-imbalance problem.

\subsubsection{Combine Different Levels of Features}
To address the varying discriminative power of each time step in $X_{TCN}$, we utilize an attention mechanism to aggregate $X_{TCN}$. The attention score is computed using $X_{I3D} \in \mathbb{R}^{d}$ extracted from the I3D Network, which contains important low-level motion features that can assist in calculating attention weights. The operation is as follows:

\begin{equation}
    X_{attn} = \text{Softmax}(\text{MLP}_1(X_{I3D}) \times X_{TCN})
\end{equation}
\begin{equation}
    X_{HL} = X_{TCN} \times X_{attn}^\text{T}
\end{equation}

where $\text{MLP}_1$ consists of two fully connected (FC) layers and a dropout layer with the last FC layer having $C$ hidden units, $X_{attn} \in \mathbb{R}^{1\times T}$ is the attention score for $X_{TCN}$, and $X_{HL} \in \mathbb{R}^{C}$.

As mentioned earlier, the features extracted by the I3D Network can capture information that may be overlooked by high-level features. Therefore, in addition to the high-level features $X_{HL}$, we combine them with $X_{I3D}$ to create the final feature representation for downstream prediction. Therefore, we concatenate the information, resulting in:
\begin{equation}
X_{vis} = \text{CONCAT}(\text{MLP}_2(X_{I3D}), X_{HL})
\end{equation}

Here, $\text{MLP}_2$ consists of two FC layers with a rectified linear unit (ReLU) layer in between. The last FC layer has $C$ hidden units. By concatenating the output of $\text{MLP}_2$ with $X_{HL}$, we obtain the final feature representation $X_{vis} \in \mathbb{R}^{2C}$. Subsequently, the model prediction based on $X_{vis}$ can be formulated as:

\begin{equation}
    s = \text{MLP}_3(\text{NORM}(X_{vis}))
\end{equation}

$\text{MLP}_3$ is implemented as a normalized FC layer which incorporates a normalized weight vector without bias. With $X_{vis}$ being normalized, $s$ is a scalar within $[-1,1]$. A higher value suggests a more engaged subject. Following \citet{kaur2018prediction} which assigned a scalar to each engagement class with a uniform gap, we employ a uniform threshold of ($-0.5, 0, 0.5$) to classify the data into one of the four engagement levels, namely highly disengaged (HD), disengaged (DE), engaged (EG), and highly engaged (HE).

Considering that engagement is not merely confined to discrete categories but exists along a spectrum, we decide to predict engagement as a scalar, which allows for a continuous representation of engagement levels and emphasizes the ordinal relationship. It enables the model to capture subtle variations and nuances in the level of engagement, which may vary within the same engagement class. 

\subsubsection{Audio Features}
In addition to the vision features, we incorporate audio features into the prediction. We utilize a pre-trained BERT model \cite{devlin2018bert}, specifically the bert-base-uncased model from HuggingFace, to extract information from the speech $sp$. Regarding the acoustics, we select metadata of the volume and the pitch. The operation is as below:
\begin{equation}
    X_{sp} = \text{FC}(\text{BERT}(sp))
\end{equation}
\begin{equation}
    X_{aud} = [L, Hv, Lv, Hp, Lp, std_v, std_p]
\end{equation}
\begin{equation}
    X^{'} = \text{CONCAT}(X_{vis}, X_{sp}, X_{aud})
\end{equation}
\begin{equation}
     s = \text{MLP}_3(\text{Norm}(X^{'}))
\end{equation}

where $X_{sp} \in \mathbb{R}^{768}$, $X_{aud} \in \mathbb{R}^7$ includes the speech length $L$, percentage of high volume $Hv$, percentage of low volume $Lv$, percentage of high pitch $Hp$, percentage of low pitch $Lp$, and standard deviation of volume $std_v$ and pitch $std_p$. These multi-modal features, denoted as $X^{'}$, are then sent into the normalized FC layer, which is identical to the process when only vision features are considered.


\subsection{MocoRank}
Subjects within the same class may exhibit diverse behaviors and display similar yet not identical engagement levels. To address intra-class variations effectively when designing the loss criteria, it is crucial to avoid imposing a common ground truth on each class. However, setting a common ground truth for each class is required when training with MSE Loss. On the other hand, training with Cross-Entropy Loss may ignore the ordinal relationship between each class. We present \textbf{MocoRank} which is specifically designed to handle the complexities of intra-class variations and ordinal relationships, enabling more accurate and robust learning for engagement prediction.

Taking inspiration from \citet{He_2020_CVPR}, we introduce MocoRank to train the model using relative assessments. Instead of relying on individual data points, comparisons between different data can facilitate better representation learning for minority classes in data imbalance situations. MocoRank consists of two parts. First, the model predicts a score for each data in the mini-batch. These scores and the score pool are used to calculate the Multi-Margin Loss to update the model. Secondly, the same batch of data is sent to the momentum encoder and the score pool is updated with the output from the momentum encoder.

\subsubsection{Momentum Encoder and Score Pool}
We use the sampling mechanism in MoCo \cite{He_2020_CVPR} to maintain the score pool which contains pre-predicted engagement scores. These scores are generated by the momentum encoder. The score pool provides a set of reference points, which is later used by Multi-Margin Loss to evaluate the suitability of newly predicted scores. 

The momentum encoder shares the same structure as the model and they are initialized with the same weights. Besides, the momentum encoder is updated for each iteration by retaining 99.9\% of its current weight and incorporating 0.1\% of the model's weight. This gradual update of the momentum encoder ensures consistent and stable generated scores, preventing excessive fluctuations between iterations.

In each iteration after the model predicts the scores for the mini-batch with size $\lvert B \lvert$, the momentum encoder processes the mini-batch and produces a score for each data. These scores, feature embeddings (feature before the MLP$_3$ of the momentum encoder), and ground truth labels constituted into triplets. These triplets are stored in the score pool, which operates on a first-in, first-out (FIFO) principle. The score pool has a predetermined length of $\lvert P \lvert$ and it is initially filled with triplets from four different engagement levels, shuffled randomly to ensure a diverse mix of examples. Similar to the rationale behind the update of the momentum encoder, the score pool is updated as a queue by replacing the $\lvert B \lvert$ most outdated data triplets in each iteration to ensure a steady transition.

\begin{table*}[ht]
\fontsize{9pt}{0pt}
\centering
\begin{tabular}{l|c|c|c|c|c|c|c|c|c|c}
\toprule
 \textbf{Loss}   & \multicolumn{2}{c|}{\textbf{Our Backbone}} & \multicolumn{2}{c|}{\textbf{GAP + GRU}} & \multicolumn{2}{c|}{\textbf{Resnet + TCN}} & \multicolumn{2}{c|}{\textbf{SlowFast}} & \multicolumn{2}{c}{\textbf{VIVIT}} \\  
   &    \textbf{Acc.} & \textbf{AvgAcc.}  &    \textbf{Acc.} & \textbf{AvgAcc.}  &    \textbf{Acc.} & \textbf{AvgAcc.}  &    \textbf{Acc.} & \textbf{AvgAcc.}  &    \textbf{Acc.} & \textbf{AvgAcc.}     \\ 
\midrule
Class Sampler+CE   &    $75.91$ & $53.41$ &$65.81$ & $44.17$ & $65.68$ &$39.78$ & $72.89$ &$53.28$& $70.43$&$51.55$  \\ 
Class Sampler+MSE   &    $76.73$ & $53.81$  & $61.58$& $50.68$& $65.76$ &$42.60$ &$74.52$&$52$&$70.10$&$54.8$  \\ 
MSE+Rank Loss       &    $76.07$ & $53.92$ & $57.45$&  $51.37$&  $69.77$ &$43.14$ &$73.71$&$52.28$&$71.25$&$56.93$   \\ 
MSE+Triplet Loss  &    $76.15$ & $53.32$ & $57.45$& $47.83$& $69.52$ &$43.56$ &$73.66$&$53.78$&$71.25$&$55.88$       \\ 
CE+Center Loss  &    $76.07$ & $55.89$ & $66.06$& $50.86$ &$68.55$ &$42.20$ 
 &$74.44$&$53.48$&\underline{$74.77$}&$55.08$      \\ 
CB Focal Loss &    $76.82$ & $54.47$ & $64.65$ & $50.90$ & $65.19$ &$37.07$ & $76.14$ &$56.51$& $71.41$&$54.46$       \\ 
 \midrule
MocoRank  &   $77.48$ & \underline{$\bf 60.94$} & $66.80$ & $51.30$ &$71.49$ &$46.02$ & $76.41$ &$57.06$ & $73.79$ &\underline{$59.06$}       \\ 
MocoRank+Center Loss  &  \underline{$\bf 78.14$}& $55.74$ & \underline{$67.60$} & \underline{$51.64$} & \underline{$72.8$}  & \underline{$47.23$} & \underline{$76.98$} & \underline{$58.83$} & $74.36$  &$56.92$        \\ 
\bottomrule
\end{tabular}
\caption{Accuracy of different architectures trained with different loss. We underline the highest
accuracy in each column and make the highest accuracy and average accuracy in the table bolded.}
\label{tab:results}
\end{table*}

\subsubsection{Multi-Margin Loss}
The Multi-Margin Loss uses the scores $S$ generated by the model, in combination with the score pool $P$, to calculate the loss for updating the weight of the model. The Multi-Margin Loss w.r.t to one batch of training samples can be formulated as:
\[  \fontsize{7.5pt}{0pt} L = \dfrac{1}{|B| \times |P|}\sum_{l_1, d_1 \in B} \sum_{l_2,s_2,e_2 \in P} \text{max}(f(l_1,d_1,l_2,s_2,e_2),0)
\]

where $B$ denotes the training batch and data $d_1$ with its label $l_1$ representing one training sample of $B$. $P$ is the score pool where each element is a triplet of ground truth label, predicted score, and embedding $(l_2,s_2, e_2)$. Specifically, $f$ is formulated as:
\[  \fontsize{7.5pt}{0pt}
    f(l_1,d_1,l_2,s_2,e_2) = 
        \begin{cases}
            L_1(\text{model}(d_1)-s_2) & \text{if } l_2 = l_1\\
            M_{\lvert l_2-l_1 \rvert}-(\text{model}(d_1)-s_2)& \text{if } l_1 > l_2\\
            M_{\lvert l_2-l_1 \rvert}-(s_2-\text{model}(d_1))& \text{if } l_1 < l_2\\
        \end{cases}
\]

The margin $M_{\lvert l_2-l_1 \rvert}$ determines the lowest engagement score difference that can be tolerated. It is determined by two factors: the difference between the labels of the two data points and the cosine similarity between the two embeddings. In detail, $M_{\lvert l_2-l_1 \rvert}$ is formulated as:

\begin{equation}
    M_1:  0.5 * (\text{CosineSimilarity}(e_1, e_2)+1)/2 
\end{equation}
\begin{equation}
    M_2:  0.5 + 0.5*(\text{CosineSimilarity}(e_1, e_2)+1)/2
\end{equation}
\begin{equation}
    M_3:  1.0 + 0.5*(\text{CosineSimilarity}(e_1, e_2)+1)/2
\end{equation}

$e_1$ is the feature embedding of $d_1$, and $e_2$ is the feature embedding of $d_2$ obtained previously from the momentum encoder and saved in the score pool. A larger margin is imposed when the label difference between the two data points is larger or when the feature embeddings of the two data points from different classes are similar. For example, suppose $d_1$ is HE ($l_1=3$) and $d_2$ is DE ($l_2=1$), then the loss will be calculated as $M_2-(\text{model}(d_1)-s_2)$, where if $M_2 > (\text{model}(d_1)-s_2)$, the model will receive penalty from the loss.

The multi-margin loss is based on the idea that a subject's engagement level should be predicted as a score higher than less engaged subjects and lower than more engaged subjects. This score difference should surpass the margin determined by the label difference. The loss highlights score relativity without requiring specific ground truth data. We employ cosine similarity for flexible threshold computation, penalizing similar representations across different classes and easing penalties for well-predicted score relativity.
\section{Experiment}

\subsection{Implementation Details} 

We use AdamW optimizer with a weight decay of 1$e$-3, batch size $|B|=256$, and score pool length $|P|=2048$ for training. The number of training epochs is 1200 with an initial learning rate of 5$e$-4 decayed to 5$e$-7 using the CosineAnnealing Scheduler. We report models' overall accuracy (Acc.) and average accuracy (Avg Acc.).

The combination of different modality features is carried out asynchronously. Initially, only the visual features are utilized to train the visual feature extractor. Subsequently, the audio features are incorporated while freezing the visual feature extractor to train the audio feature extractor. The reason behind this separate training approach for multi-modality input is based on our observation that simultaneously training with both types of features leads to a decline in performance. We suspect that the imbalance in the quantity of visual and audio inputs is the cause, as participants do not speak continuously throughout the class.

\subsection{Main Results}

In Table \ref{tab:results}, we compare the performance of MocoRank with loss functions and architectures proposed by previous studies. The compared losses include LDAM \cite{cao2019learning}, Center Loss \cite{liao2021deep}, Rank Loss \cite{wang2019bootstrap}, and Triplet Loss \cite{copur2022engagement}. The weights for Center Loss, Rank Loss, and Triplet Loss are set to $0.2$, $1$, and $1$, following the original setting specified in their papers. For architectures, we compare our model with previous work \cite{abedi2021improving, niu2018automatic}, SlowFast \cite{feichtenhofer2019slowfast}, and VIVIT \cite{arnab2021vivit}. 

Overall, we can observe that MocoRank outperforms the other loss functions in both accuracy and average accuracy across all architectures. Compared with CE+Center Loss, an improvement of 5.05\% in average accuracy suggests that MocoRank can better handle the imbalance setting. Furthermore, when we incorporate Center Loss into MocoRank, we achieve an even higher accuracy of 78.14\%. 

\begin{figure}
\centering
\includegraphics[scale=0.35]{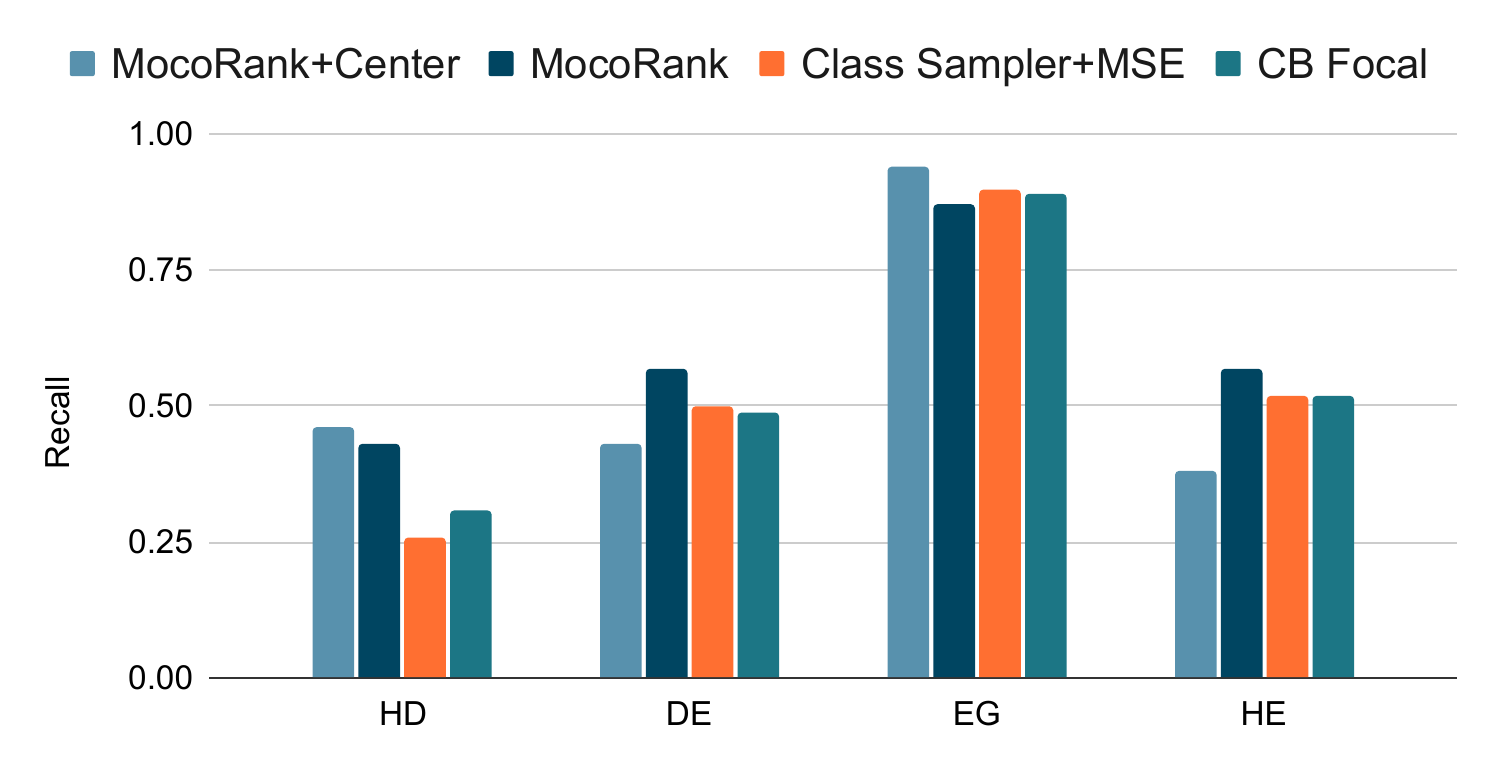}
\caption{A comparison of model recall on each class using differ- ent loss for training.}
\label{fig:lossdis}
\end{figure}

Figure \ref{fig:lossdis} illustrates the distribution of recalls among the four loss functions that yield the highest results. MocoRank performs significantly better than the others in the HD, DE, and HE classes, and competitive performance in the EG class. The superior performance of MocoRank shows that it can learn the features of minority classes more effectively.

\subsection{Ablation Studies}
\subsubsection{Model Architecture}
We examine and exclude several branches in the method we proposed to combine high-level features and I3D features. In Table \ref{tab:attconcat}, the result suggests that using I3D features to provide attention can improve the accuracy by 3.65\%, and using concatenation of the two features can improve by 6.54\%. By combining the two methods, the accuracy can further be improved by 7.7\%. The result suggests that our model design is beneficial to representation learning. 

We examine different temporal modules in our model and train the models with MocoRank. From Table \ref{tab:bone}, we explore that TCN outperforms other intricate modules like Transformer or Bi-LSTM.

\begin{table}
\fontsize{9pt}{0pt}
\centering
\begin{tabular}{l|c|c}
\toprule
\textbf{Method}   & \textbf{Acc.(\%)} & \textbf{Avg Acc.(\%)} \\ 
\midrule
Only Openface & $70.44$ & $46.76$ \\
Attention    &    $74.09$           &     $50.32$\\ 
Concat            &    $76.98$      &      $52.50$ \\ 
Concat+Attention       &    $\bf 78.14$ &      $\bf 55.74$     \\ 
\bottomrule
\end{tabular}
\caption{Accuracy of different methods to combine high-level features and I3D features.}
\label{tab:attconcat}
\end{table}

\begin{table}
\fontsize{9pt}{0pt}
\centering
\begin{tabular}{l|c|c}
\toprule
\textbf{Feature}   & \textbf{Acc.(\%)} & \textbf{Avg Acc.(\%)} \\ 
 \midrule
Transformer      &    $75.99$ & $54.07$          \\ 
Bi-LSTM       &    $75.74$ & $56.73$          \\ 
TCN      &    $\bf 77.48$ & $\bf 60.94$          \\ 
\bottomrule
\end{tabular}
\caption{Accuracy of different temporal networks.}
\label{tab:bone}
\end{table}

\subsubsection{Multi-Modality Features}
In Table \ref{tab:featureacc}, we evaluate the performance of a model trained on different feature combinations. When utilizing both visual and audio features, we employ only the visual segment to make predictions for video segments without speech, and the full model is utilized for predicting video segments that include audio and speech. We can observe that incorporating multi-modalities can improve the performance. To further examine the effect of adding audio features, we evaluate the model on the test subset which consists of data with speech. The last two rows in Table \ref{tab:featureacc} show that audio features can increase the accuracy by 3.18\% and the average accuracy by 3.47\%, which suggests audio features could add information complementary to visual features.

\begin{table}
\fontsize{9pt}{0pt}
\centering
\begin{tabular}{l|c|c}
\toprule
\textbf{Feature}   & \textbf{Acc.(\%)} & \textbf{Avg Acc.(\%)} \\  
\midrule
Openface       &    $70.44$ & $46.76$          \\ 
I3D       &    $72.43$ & $58.76$          \\ 
Openface+I3D     &    $78.14$ & $55.74$            \\ 
Openface+I3D+Audio   &    $\bf 78.55$ & $\bf 56.85$  \\ 
\midrule
Openface+I3D     &    $71.42$ & $60.23$          \\ 
Openface+I3D+Audio     &    $\bf 74.60$ & $\bf 63.70$            \\ 
\bottomrule
\end{tabular}
\caption{Accuracy of the model using different combinations of features. The first four rows are evaluated in the full test set. The bottom two rows are evaluated on the test subset consisting of data with audio and speech features.}
\label{tab:featureacc}
\end{table}

\subsection{Data Transferability}
We use the same setting to train the model on EngageWild and DAiSEE. Next, for each model trained on one dataset, we finetune the model with the other two datasets for 250 epochs using MocoRank and Center Loss.
 
Table \ref{tab:transfer} shows that after finetuning, models pre-trained on the CMOSE dataset outperform models pre-trained on EngageWild and DAiSEE. For instance, when evaluating performance on EngageWild, the model pre-trained on CMOSE can achieve 6.25\%  higher accuracy than the model trained on EngageWild itself. A similar result is shown when evaluating performance on DAiSEE, where the model pre-trained on CMOSE has a 2.36\% improvement compared to the model trained on DAiSEE. Notably, the incompatible performance suggests neither EngageWild nor DAiSEE features transfer effectively to CMOSE. This outcome underscores CMOSE's feature transferability superiority relative to other engagement datasets.

\begin{table}
\fontsize{9pt}{0pt}
\centering
\begin{tabular}{l|c|c|c}
\toprule
 \textbf{Dataset}   & \textbf{EngageWild} & \textbf{DAiSEE} & \textbf{CMOSE}\\ 
 \midrule
   EngageWild        &    $45.83$     &    $49.32$   & $52.15$ \\ 
   DAiSEE    &    $35.41$     &    $47.70$ & $51.40$ \\ 
    CMOSE   &  $\bf 52.08$    &    $\bf 51.68$   & $\bf 78.14$ \\ 
\bottomrule
\end{tabular}
\caption{Comparison of transferability. The column indicated the dataset used to fine-tune and evaluate the model. The row indicates the dataset the model is pre-trained on.}
\label{tab:transfer}
\end{table}

\section{Conclusion}
With the surge in online classes, engagement prediction has gained significant attention. This paper advances engagement prediction across three dimensions. First, we present the CMOSE dataset, which contains sufficient data at each engagement level and high-quality labels based on the psychologist's advice. Secondly, we propose MocoRank to alleviate the data imbalance problem. Lastly, we show that the fusion of vision and audio features can improve performance in engagement prediction.  

While our results show a promising direction in engagement prediction, there is more to be explored. First, the CMOSE dataset involves 102 participants and a future direction could be to personalize the model prediction. Another possible direction is to explore certain behaviors from the coaches that may increase or decrease the engagement level of the students.

\textbf{Acknowledgment:} This work, the dataset construction, and the annotation process are supported by LifeHikes.

\small
\bibliographystyle{ieeenat_fullname}
\bibliography{main}

\clearpage
\section{Supplementary Materials}
In this section, we list out the shape of all the input features of the model. The detail is listed in Table \ref{tab:featuredim}.

\begin{table}[h]
\small
\centering
\begin{tabular}{l|l}
\toprule
Feature   & Shape \\ \midrule
Gaze      & $(B \times 8)$  \\ 
Head Pose   & $(B \times 6)$           \\ 
Facial Action Units & $(B \times 35)$              \\ \midrule
I3D vector   & $(B \times 1024)$              \\ \midrule
Speech      & String     \\ \midrule
   & Volume: $(B \times n)$ \\ 
Acoustics  & Pitch: $(B \times m)$ \\ 
 & n,m depends on speech length \\ 
\bottomrule
\end{tabular}
\caption{Format of different extracted features}
\label{tab:featuredim}
\end{table}

We invited two psychology experts to suggest and evaluate the importance of certain engagement-related behaviors. A lower score from the expert means that the behavior may indicate disengaging. In contrast, a higher score means the behavior may be engaging. The detail is listed in Table \ref{tab:expert}.

\begin{table}[h]
\footnotesize
\centering
\begin{tabular}{l|c|c|c}
\toprule
Feature   & Expert1 & Expert2 & Type  \\
\midrule
Arms crossed & 4 & 7 & Body \\
Consistent pose & 5 & 6 & Body \\
Changing seating position & 6 & 6 & Body \\
Slouching & 3 & 4 & Body \\
Sudden behavior change & 8 & 5 & Body \\
Yawning & 3 & 5 & Body \\
\midrule
Back from breaking room & 8 & 10 & Facial \\
Speaking & 6 & 10 & Facial \\
Smile & 7 & 7 & Facial \\
\midrule
Active hand movements & 8 & 7 & Hand \\
Hand at the back of head & 3 & 4 & Hand \\
Drinking or eating & 5 & 6 & Hand \\
Gesture+Speaking & 7 & 7 & Hand \\
Playing hands & 3 & 5 & Hand \\
Hand on mouth (thinking) & 6 & 8 & Hand \\
Hand stretching & 5 & 6 & Hand \\
Modify glasses & None & 5 & Hand \\
\midrule
Moving closer to screen & 8 & 9 & Head \\
Nodding & 9 & 10 & Head \\
Head tilting towards screen & 4 & 6 & Head \\
\midrule
Looking down & 4 & 4 & Gaze \\
Blank stare & 3 & 4 & Gaze \\
Eye rolling & 4 & 5 & Gaze \\
Focus on other objects & 3 & 5 & Gaze \\
Focus on a point on screen & 5 & 7 & Gaze \\
Consistent gaze direction & 5 & 6 & Gaze \\
\bottomrule
\end{tabular}
\caption{Different behaviors and their received scores from psychology experts.}
\label{tab:expert}
\end{table}

\end{document}